\pdfoutput=1
\documentclass[11pt]{article}

\usepackage{ACL2023}

\usepackage{times}
\usepackage{latexsym}

\usepackage[T1]{fontenc}
\usepackage[utf8]{inputenc}

\usepackage{graphicx}
\usepackage{subfigure}
\usepackage{float}
\usepackage{amsmath}
\usepackage{amssymb}
\usepackage{multirow}
\usepackage{booktabs}
\usepackage{url}
\usepackage{array}
\usepackage{enumitem}
\usepackage{algorithm}
\usepackage{algorithmic}
\usepackage{pifont}
\usepackage{bm}
\usepackage{lipsum}
\usepackage{xcolor}
\usepackage{makecell}

\setenumerate[1]{itemsep=5pt,partopsep=0pt,parsep=\parskip,topsep=5pt}
\setitemize[1]{itemsep=5pt,partopsep=0pt,parsep=\parskip,topsep=5pt}
\setdescription{itemsep=5pt,partopsep=0pt,parsep=\parskip,topsep=5pt}

\usepackage{microtype}

\usepackage{inconsolata}

\title{LERT: A Linguistically-motivated Pre-trained Language Model}

\author{Yiming Cui$^{1,2,\dag}$, Wanxiang Che$^1$, Shijin Wang$^{2,3}$, Ting Liu$^1$ \\
{$^1$Research Center for SCIR, Harbin Institute of Technology, Harbin, China} \\
{$^2$State Key Laboratory of Cognitive Intelligence, iFLYTEK Research, Beijing, China} \\
{$^3$iFLYTEK AI Research (Central China), Wuhan, China} \\
{\tt $^\dag$ymcui@ieee.org} \\  
}

\begin{document}
\maketitle
\begin{abstract}
Pre-trained Language Model (PLM) has become a representative foundation model in the natural language processing field.
Most PLMs are trained with linguistic-agnostic pre-training tasks on the surface form of the text, such as the masked language model (MLM).
To further empower the PLMs with richer linguistic features, in this paper, we aim to propose a simple but effective way to learn linguistic features for pre-trained language models.
We propose LERT, a pre-trained language model that is trained on three types of linguistic features along with the original MLM pre-training task, using a linguistically-informed pre-training (LIP) strategy.
We carried out extensive experiments on ten Chinese NLU tasks, and the experimental results show that LERT could bring significant improvements over various comparable baselines.
Furthermore, we also conduct analytical experiments in various linguistic aspects, and the results prove that the design of LERT is valid and effective.\footnote{Pre-print version, subjected to changes. Resources are available at \url{https://github.com/ymcui/LERT}}
\end{abstract}

%%%%%%%%%%%%%%%%%%%%%%%%%%%%%%%%%%%%%%%%
\section{Introduction}
Pre-trained Language Model (PLM) has been proven to be a successful way for text representation, which considers rich contextual information.
Among several types of pre-trained language models, auto-encoding PLMs, such as BERT \citep{devlin-etal-2019-bert} and RoBERTa \citep{liu2019roberta}, are relatively popular for natural language understanding (NLU) tasks.
Unlike the auto-regressive PLMs (e.g., GPT \citep{radford2018improving}) that use a standard language model as the training objective, auto-encoding PLMs largely rely on pre-training tasks to learn contextual information.
Masked language model (MLM), which was first proposed in BERT, has been a dominant pre-training task for auto-encoding PLMs, such as RoBERTa, ALBERT \citep{lan2019albert}, ERNIE \citep{sun2019ernie}, DeBERTa \citep{he2021deberta}, etc., demonstrating its broad generalizability in learning text representations.
The MLM task learns to recover word information from the masked text, where the masked word is usually chosen randomly, indicating that MLM is a linguistic-agnostic pre-training task without explicit utilization of linguistic knowledge.

Though it is widely perceived that the pre-trained language model entails rich linguistic knowledge \citep{jawahar-etal-2019-bert}, some researchers propose to further include external knowledge in PLMs.
Specifically, to incorporate linguistic knowledge into the pre-trained language model, various efforts have been made in the community,
such as incorporating structural knowledge \citep{zhou-etal-2020-limit,xu-etal-2021-syntax}, including additional linguistic tasks \citep{zhang2021mengzi}, etc.
Though various efforts have been made, the previous work has several limitations.
Most of these works only focus on including several linguistic features in PLM without carefully analyzing how individual features contribute to the overall performance and the relations between different tasks. 
Also, the implementations are relatively complex, as structural knowledge cannot be directly applied into PLMs.

To alleviate the issues above, in this paper, we leverage the traditional natural language processing method to explicitly include more linguistic knowledge, creating weakly-supervised data for model pre-training.
Also, to investigate whether pre-trained language models can benefit from explicitly injecting linguistic knowledge, in this paper, we propose a new pre-trained language model called {\bf LERT} ({\bf L}inguistically-motivated bidirectional {\bf E}ncoder {\bf R}epresentation from {\bf T}ransformer).
LERT is trained on the masked language model as well as three types of linguistic tasks, including part-of-speech (POS) tagging, named entity recognition (NER), and dependency parsing (DEP), forming a multi-task pre-training scheme.
Furthermore, to balance the learning speed for each pre-training task, we propose a linguistically-informed pre-training (LIP) strategy, which learns fundamental linguistic knowledge faster than the high-level ones.
With thorough ablations and analyses, LERT has proven effective on various natural language understanding tasks over comparable baselines.
The contributions of this paper are listed as follows.
\begin{itemize}
	\item We propose a simple way to incorporate three types of linguistic features for pre-trained language models with a linguistically-informed pre-training (LIP) strategy.
	\item With extensive and robust experiments on ten popular Chinese natural language understanding tasks, LERT yields significant improvements over comparable baselines. Several analyses also prove the effectiveness of LERT.
	\item The resources are made publicly available to further facilitate our research community.
\end{itemize}

%%%%%%%%%%%%%%%%%%%%%%%%%%%%%%%%%%%%%%%%
\section{Related Work}
The recent advancement of natural language processing largely owes to the development of text representations.
Speech signals can be represented by waves, and images can be represented by pixels, where they all have clear physical concepts and can be directly represented in computers.
However, when it comes to natural language, it has no exact representation for a specific semantic.
Thus, a major research topic in NLP is to find a better way for text representation.
In the last decade, static word embedding has been a dominant text representation method in NLP, such as word2vec \citep{mikolov-etal-2013} and GloVe \citep{pennington-etal-2014-glove}.
However, these representations cannot deal with the problem of polysemy.
Later, ELMo \citep{peters-etal-2018-deep} was proposed to solve this issue, which models the text in recurrent neural networks, and word representation can be adjusted by its context.
Transformer-based \citep{vaswani2017attention} neural networks have proven effective among various NLP tasks.
A combination of the transformer model and text representation has led to the recent emergence of pre-trained language models.
Pre-trained language model (PLM), such as BERT and GPT, uses deep transformer models to encode the text in a contextual way, which can be applied to a wide range of natural language processing tasks.
The training of PLM only requires large-scale unlabeled text with self-supervised tasks, such as the masked language model.
Though linguistic knowledge is not explicitly injected, various PLMs achieve significant improvements on many NLP tasks.

One of the main reasons that make PLMs successful is that pre-trained language models learn better text semantics and entail linguistic knowledge, though they are not explicitly learned in the self-supervised task, which is commonly perceived by the community.
For example,
\citet{jawahar-etal-2019-bert} discovered that the intermediate layers of BERT capture rich linguistic information.
\citet{kovaleva-etal-2019-revealing} focuses on the multi-head self-attention mechanism itself to demonstrate its redundancies.
\citet{liu-etal-2019-linguistic} investigated the transferability of contextual representations with several linguistic probing tasks.
\citet{hewitt-manning-2019-structural} propose a structural probe for finding syntax information in pre-trained language models.
These works have brought us a better understanding of which types of linguistic features are learned in PLMs.

Some researchers also tried incorporating linguistic features in pre-trained language models to further improve their performance on downstream tasks.
\citet{zhou-etal-2020-limit} propose LIMIT-BERT, which incorporates five linguistic tasks: part-of-speech, constituent and dependency parsing, span, and dependency semantic role labeling (SRL).
\citet{xu-etal-2021-syntax} propose a syntax-enhanced pre-trained model, which incorporates a syntax-aware attention layer during both the pre-training and fine-tuning stages.
\citet{zhang2021mengzi} utilizes part-of-speech tagging and named entity recognition as additional linguistic tasks during pre-training.
\citet{liu-etal-2021-lexicon} propose LEBERT for Chinese sequence labeling, which incorporates external knowledge into BERT layers.
\citet{zhang2022revisiting} propose CKBERT, which uses linguistic-aware MLM and contrastive multi-hop relation model for pre-training.

\begin{figure*}[htp]
  \centering
  \includegraphics[width=1\textwidth]{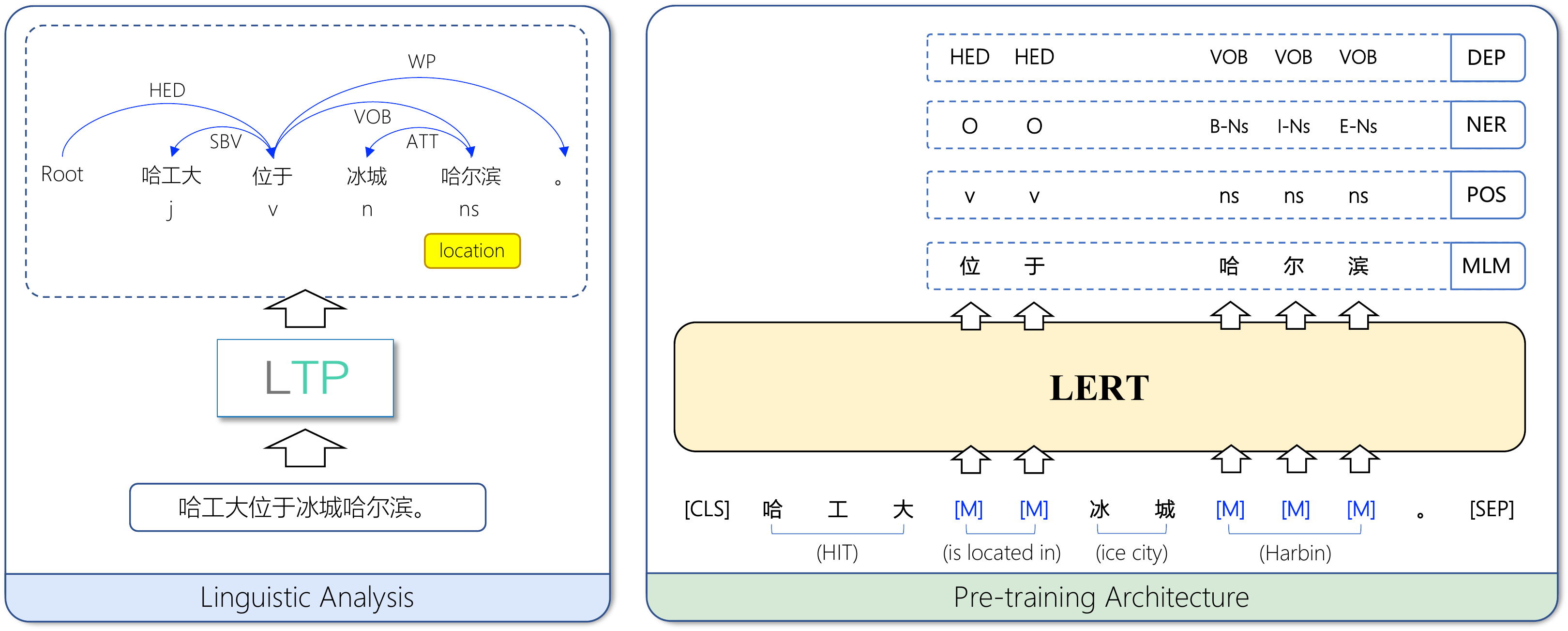}
  \caption{\label{lert-overview} Overview of LERT. We only use one text segment in the input for simplicity, i.e., we still use two text segments for implementation, separated by {\tt [SEP]} token. } 
\end{figure*}

Unlike previous works that either depend on incorporating structural knowledge with complex model design or without a clear exposition of the contribution by each linguistic feature, in this paper, we proposed LERT, which aims to directly utilize the linguistic tags for multi-task pre-training.
The pre-training task is chosen by careful analysis, and the LERT also benefits from a linguistically-informed pre-training scheme, which is in line with intuitive thinking.
The experiments and analyses present a clear contribution of each linguistic feature as well as other components to explicitly allow us to understand which type of linguistic feature is the most helpful in creating a better pre-trained language model.

%%%%%%%%%%%%%%%%%%%%%%%%%%%%%%%%%%%%%%%%
\section{LERT}

%%%%%%%%%%%%%%
\subsection{Overview}
An overview of the proposed LERT is depicted in Figure \ref{lert-overview}.
The formulation of LERT is simple and straightforward.
Firstly, we perform linguistic analysis on the given input text to get word segmentation information and extract its linguistic features.
The word segmentation information is used to perform Chinese whole word masking (wwm) and N-gram masking \citep{chinese-bert-wwm} (identical to MacBERT \citep{chinese-bert-wwm}, PERT \citep{cui2022pert}, etc.) in masked language model task.
The linguistic features are used for linguistic task pre-training.
Then we use the extracted linguistic features to perform multi-task pre-training along with the original MLM task, which linguistically-informed pre-training scheme.

%%%%%%%%%%%%%%
\subsection{Linguistic Features}
In this paper, we aim to utilize linguistic features in a simple way.
To meet this criterion, the generated linguistic feature should have two characteristics: high-accuracy and uniqueness.
High-accuracy means that the linguistic feature should be highly reliable in terms of tagging performance.
Though current language analysis tools, such as LTP \citep{che2010ltp}, Stanford CoreNLP \citep{manning-etal-2014-stanford}, etc., are capable of analyzing the linguistic features for the text, not all of the features are highly accurate.
Uniqueness means that each input token should have exactly one target tag for a specific linguistic feature, whereas most tree-based or graph-based linguistic analyses do not meet this standard and requires further complex processing.

Considering both conditions, in this paper, we use LTP \citep{che2010ltp} for annotating linguistic tags for the input text with three types of linguistic features, i.e., part-of-speech (POS), named entity recognition (NER), and dependency parsing (DEP).
These three types of linguistic features are relatively fundamental and achieve good performance on tagging and meet one-to-one tagging conditions.\footnote{POS: 98.4\% (P), NER: 91.7\% (F), DEP: 84.8\ (UAS), according to \url{http://ltp.ai/docs/ltp3.x/theory.html}}
A complete list of linguistic tags are depicted in Table \ref{linguistic-tags}.
Specifically, 
\begin{itemize}
	\item {\bf POS}: Each input token is assigned to a unique POS tag, resulting in 28 types.
	\item {\bf NER}: We use the ``BIEOS'' tagging scheme to annotate input tokens with NER information, resulting in 13 types. 
	\item {\bf DEP}: We perform syntactic dependency parsing on the input sequence. Note that we attribute the relation label to its dependent (not head) to ensure each token has a unique label, resulting in 14 types. 
\end{itemize}

After getting these linguistic labels for each input token, we can treat them as weakly-supervised labels for pre-training, where we illustrate pre-training tasks in the next.

\begin{table}[h]
\small
\begin{center}
\begin{tabular}{c p{5.7cm}}
\toprule
\bf Type (\#) & \bf Tags (abbreviation)  \\
\midrule
\multirow{11}{*}{\makecell[ct]{\bf POS \\ (28)}}  	& noun (n), verb (v), punctuation (wp), auxiliary (u), adverb (d), adjective (a), number (m), preposition (p), pronoun (r), geographical name (ns), conjunction (c), quantity (q), temporal noun (nt), person name (nh), direction noun (nd), abbreviation (j), idiom (i), other noun-modifier (b), organization name (ni), other proper noun (nz), location noun (nl), descriptive words (z), suffix (k), foreign words (ws), onomatopoeia (o), prefix (h), exclamation (e), non-lexeme (x) \\ 
\midrule
\multirow{3}{*}{\makecell[ct]{\bf NER \\ (13)}} 	& outside (O), single (S-Ni/Ns/Nh), organization names (B/I/E-Ni), person names (B/I/E-Nh), location names (B/I/E-Ns) \\
\midrule
\multirow{7}{*}{\makecell[ct]{\bf DEP \\ (14)}} 	& attribute (ATT), punctuation (WP), adverbial (ADV), verb-object (VOB), subject-verb (SBV), coordinate (COO), right adjunct (RAD), head (HED), preposition-object (POB), complement (CMP), left adjunct (LAD), fronting-object (FOB), double (DBL), indirect-object (IOB) \\
\bottomrule
\end{tabular}
\end{center}
\caption{\label{linguistic-tags} A list of linguistic tags used in LERT.}
\end{table}

%%%%%%%%%%%%%%
\subsection{Model Pre-training}
LERT is trained on the masked language model as well as three linguistic tasks, forming a multi-task training scheme.

\subsubsection{MLM Task}
For the MLM task, we follow most of the previous works that only make predictions on the masked positions\footnote{Note that ``masked positions'' includes three types of masking: ``replace with [MASK]'', ``keep original word'', and ``replace with the random word''.} instead of the whole input sequence.
We denote the last hidden layer representation of $L$-layer transformer as $\bm{H} \in \mathbb{R}^{N \times d}$ ($N$ is the length of input sequence, $d$ is hidden size), and a subset of representations w.r.t. masked positions as $\bm{H}^m \in \mathbb{R}^{k \times d}$ ($k$ is the number of masked positions), where $\bm{H}^m \subset \bm{H}$.
Then we use a fully-connected layer, followed by a layer normalization layer on $\bm{H}^m$.
\begin{equation} 
	\tilde{\bm{H}}^{m} = \mathbf{LayerNorm}(\mathbf{FFN}(\bm{H}^{m}) )
\end{equation}

We use the input word embedding matrix $\bm{E} \in \mathbb{R}^{V \times d}$ ($V$ is the vocabulary size) to project the $\tilde{\bm{H}}^{m}$ into vocabulary space, and use the softmax function to get normalized probabilities. 
\begin{equation}
	\bm{p}_i = \mathbf{softmax}( \tilde{\bm{H}}^\text{m}_i \bm{E}^\top + \bm{b} ), ~~\bm{p}_i \in \mathbb{R}^{V}
\end{equation}

Finally, we use the standard cross-entropy loss to optimize the MLM pre-training task.
\begin{equation}\label{equation-ce-loss}
	\mathcal{L}_\text{MLM} = -\frac{1}{M}\sum_{i=1}^M \bm{y}_i \log \bm{p}_i
\end{equation}

\subsubsection{Linguistic Tasks}
For each linguistic task, we treat it as a classification task.
Each input token is projected to its linguistic feature (POS, NER, and DEP), which was annotated using the method described in the previous section.
Specifically, given the representation $\tilde{\bm{H}}^{m}$, we use a fully-connected layer to project it into linguistic labels for each task.
\begin{equation}\label{eq-linguistic-task}
	\bm{p}^{\star}_i = \mathbf{softmax}( \tilde{\bm{H}}^\text{m}_i {\bm{W}^{\star}}^\top + \bm{b}^{\star} ), ~~\bm{p}_i \in \mathbb{R}^{V^{\star}}
\end{equation}

In Equation \ref{eq-linguistic-task}, the $\star$ can be one of three linguistic tasks, and $V^\star$ denotes the number of linguistic labels for each task.
We use standard cross-entropy loss to optimize each linguistic task.

\subsubsection{Linguistically-informed Pre-training}
Finally, the overall training loss is formulated as follows, where $\lambda_i \in [0, 1]$ is the scaling factor to the respective loss $\mathcal{L}_i$ for each linguistic task (POS, NER, and DEP).
\begin{equation}\label{eq-overall-loss}
	\mathcal{L} = \mathcal{L}_\text{MLM} + \sum_i{\lambda_i \mathcal{L}_i}, ~~i \in \{\text{P}, \text{N}, \text{D}\}
\end{equation}

A vanilla pre-training scheme is to treat all subtasks as equal, resulting in the following equation.
\begin{equation}\label{eq-overall-loss-vanilla}
	\mathcal{L} = \mathcal{L}_\text{MLM} 
				+ \mathcal{L}_\text{P} 
				+ \mathcal{L}_\text{N} 
				+ \mathcal{L}_\text{D} 
\end{equation}

Intuitively, the MLM task is the most important one among all subtasks.
However, how do we decide the scaling factor $\lambda$ for each linguistic task?

In this paper, we propose a linguistically-informed pre-training (LIP) strategy to tackle this issue.
By looking into these linguistic features, they are not completely equivalent.
The NER feature depends on the output of POS tagging, while the DEP feature depends on both POS and NER tagging.
We conjecture that POS is the most fundamental linguistic feature, followed by NER and DEP.
In light of their dependencies, we assign different learning speeds for each linguistic feature, yielding faster learning of POS than NER and DEP.
This is similar to human learning, where we usually learn basic things first and then the dependent high-level knowledge.

Formally, the loss scaling parameters are determined by the current training step $t$ and constant end step for scaling $T_\star$ that control the learning speed for each linguistic task.
\begin{equation}
	\lambda_\star = \min\{\frac{t}{T_\star}, 1\}, \star \in \{\text{P}, \text{N}, \text{D}\}
\end{equation}

Specifically, in this paper, we set $T_\star$ as 1/6, 1/3, and 1/2 of the total training steps for POS, NER, and DEP features, respectively.
After 1/2 of the total training steps, the training loss will become Equation \ref{eq-overall-loss-vanilla}, where all tasks contribute equally to the overall loss.
In this way, the POS features learn the most quickly, followed by NER and DEP.
We empirically find this strategy yields better performance, and detailed analysis also proves our strategy is effective (Section \ref{sec-task-order}) and in line with intuitive thoughts.

\begin{figure}[htp]
  \centering
  \includegraphics[width=1\columnwidth]{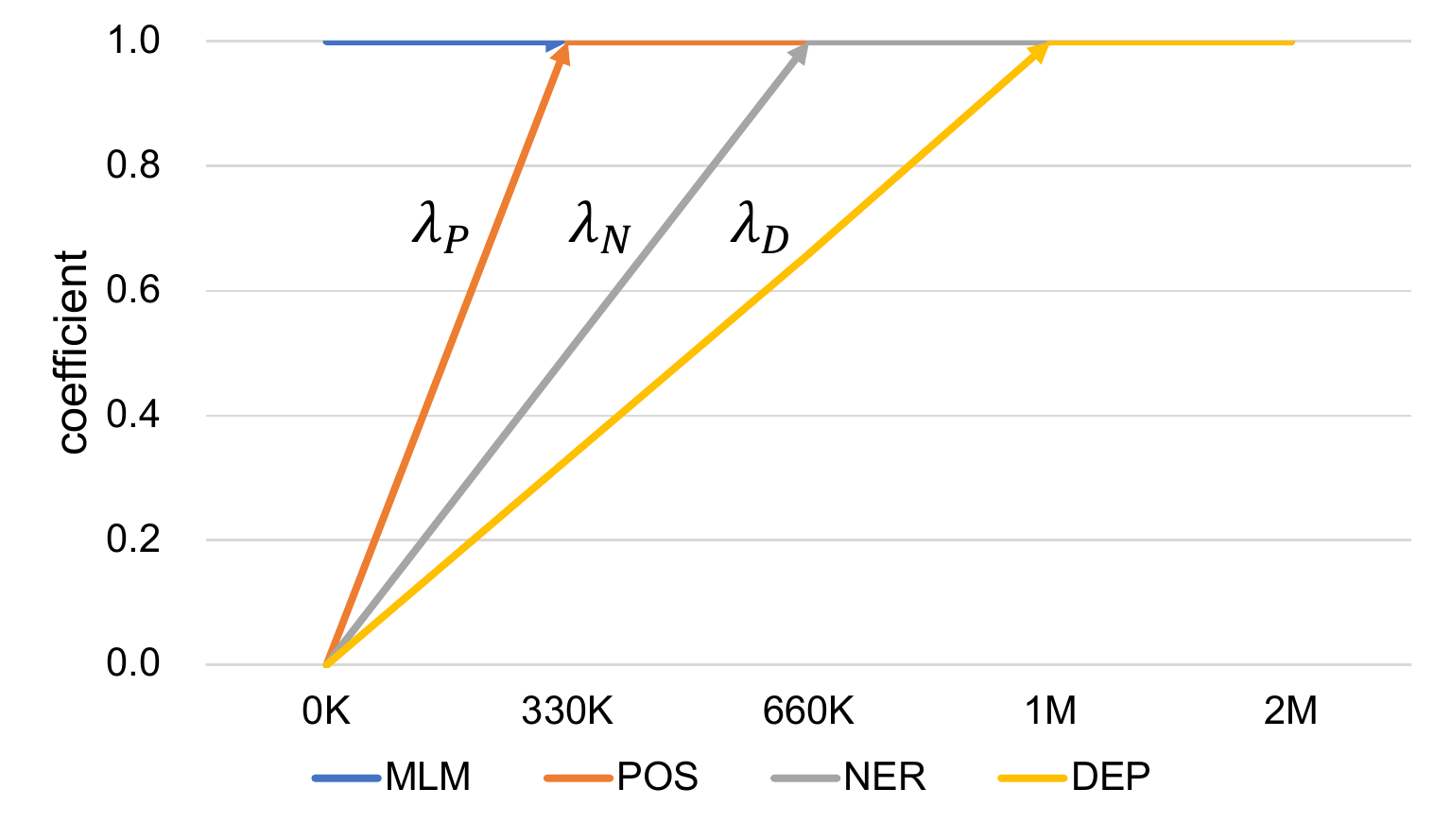}
  \caption{\label{task-warmup} Linguistically-informed pre-training strategy for LERT (with a total pre-training step of 2M).  } 
\end{figure}

%%%%%%%%%%%%%%%%%%%%%%%%%%%%%%%%%%%%%%%%
\section{Experiments}

%%%%%%%%%%%%%%
\subsection{Setups for Pre-training}

In this paper, we mainly train three LERT models.
The basic information is listed in Tabel \ref{lert-variants}.
Other pre-training setups are illustrated as follows.
\begin{itemize}[leftmargin=*]
	\item {\bf Data}: We use the training data as in MacBERT and PERT. It consists of the Chinese Wikipedia dump, encyclopedia, community question answering, news articles, etc., resulting in 5.4B words and taking about 20G of disk space.
	\item {\bf Text processing}: We use WordPiece tokenizer \citep{wu2016google} as in BERT and similar variants. All linguistic processing (such as word segmentation, tagging, etc.) is done with LTP \citep{che2010ltp}. We directly use the same vocabulary as in Chinese BERT-base with 21,128 entries.
	\item {\bf Optimization}: We use \textsc{Adam} \citep{kingma2014adam} with weight decay (rate = 0.1) optimizer using an initial learning rate of 1e-4. Each model is trained on 2M steps with the first 10K steps of linear warmup for learning rate. All models are trained from scratch.
	\item {\bf Others}: The maximum sequence length is 512. The overall masking ratio is set as 15\%.
	\item {\bf Training device}: All models are trained on a single Cloud TPU v3-8 (128G HBM) with gradient accumulation (if necessary). 
\end{itemize}
\begin{table}[h]
\small
\begin{center}
\begin{tabular}{l c c c c c}
\toprule
\bf Model & \bf Params & \bf Layers & \bf Hid. & \bf A.H. & \bf Batch  \\
\midrule
LERT$_\text{small}$ 		& 15M  & 12 & 256 & 4 & 1024 \\
LERT$_\text{base}$  		& 102M & 12 & 768 & 12 & 416 \\
LERT$_\text{large}$ 		& 325M & 24 & 1024 & 16 & 256 \\
\bottomrule
\end{tabular}
\end{center}
\caption{\label{lert-variants} Model structure for different sizes of LERT. Hid: hidden size, A.H.: attention heads.}
\end{table}

%%%%%%%%%%%%%%
\subsection{Setups for Fine-tuning Tasks}
Following previous works \citep{chinese-bert-wwm,cui2022pert}, we examine LERT's performance on ten natural language understanding tasks, including machine reading comprehension (MRC), text classification (TC), named entity recognition (NER), etc. 
Specifically,
\begin{itemize}[leftmargin=*]
	\item {\bf MRC} (2): CMRC 2018 \citep{cui-emnlp2019-cmrc2018}, DRCD \citep{shao2018drcd}.
	\item {\bf TC} (6): XNLI \citep{conneau2018xnli}, LCQMC \citep{liu2018lcqmc}, BQ Corpus \citep{chen-etal-2018-bq}, ChnSentiCorp \citep{tan2008empirical}, TNEWS \citep{clue}, OCNLI \citep{ocnli}. 
	\item {\bf NER} (2): MSRA-NER (SIGHAN 2006) \citep{levow-2006-third}, People's Daily (PD)\footnote{\url{https://github.com/ProHiryu/bert-chinese-ner}}.
\end{itemize}

\begin{table}[h]
\small
\begin{center}
\begin{tabular}{l | c c | c c c}
\toprule
\bf Dataset & \bf MaxL & \bf Ep. & \bf Train & \bf Dev & \bf Test  \\
\midrule
CMRC-18 		& 512 & 3/2/1 & 10K & 3.2K & 4.9K  \\
DRCD 		 	& 512 & 5/2/3 & 27K & 3.5K & 3.5K  \\
\midrule
XNLI 			& 128 & 5/2/2 & 392K & 2.5K & 5K  \\
LCQMC 		 	& 128 & 5/3/3 & 240K & 8.8K & 12.5K  \\
BQ Corpus	  	& 128 & 5/3/2 & 100K & 10K & 10K  \\
CSC 	& 256 & 5 & 9.6K & 1.2K & 1.2K \\
TNEWS			& 128 & 5 & 53.3K & 10K & 10K \\
OCNLI 		 	& 128 & 5/5/3 & 56K & 3K & 3K \\
\midrule
MSRA  		& 256 & 10 & 45K & - & 3.4K \\
PD	& 256 & 10 & 51K & 4.6K & - \\
\bottomrule
\end{tabular}
\end{center}
\caption{\label{ft-hyper} Hyper-parameter settings and data statistics for fine-tuning tasks. MaxL: sequence max length, Ep: training epochs (small/base/large).}
\end{table}

\begin{table*}[ht]
\tiny
\begin{center}
\begin{tabular}{l c c c c c c | c c c c}
\toprule
\multirow{2}*{\bf System} & \multicolumn{6}{c}{\centering \bf CMRC 2018} & \multicolumn{4}{c}{\centering \bf DRCD} \\
& \multicolumn{2}{c}{\centering Dev (EM/F1)} & \multicolumn{2}{c}{\centering Test (EM/F1)} & \multicolumn{2}{c}{\centering Challenge (EM/F1)} & \multicolumn{2}{c}{\centering Dev (EM/F1)} & \multicolumn{2}{c}{\centering Test (EM/F1)}  \\
\midrule
BERT$_\text{base}$ & 67.1 \tiny(65.6) & 85.7 \tiny(85.0) & 71.4 \tiny(70.0) & 87.7 \tiny(87.0) & 24.0 \tiny(20.0) & 47.3 \tiny(44.6) & 85.0 \tiny(84.5) & 91.2 \tiny(90.9) & 83.6 \tiny(83.0) & 90.4 \tiny(89.9)  \\ % BERT-wwm-ext	
RoBERTa$_\text{base}$ & 67.4 \tiny(66.5) & 87.2 \tiny(86.5) & 72.6 \tiny(71.4) & 89.4 \tiny(88.8) & 26.2 \tiny(24.6) & 51.0 \tiny(49.1) & 86.6 \tiny(85.9) & 92.5 \tiny(92.2) & 85.6 \tiny(85.2) & 92.0 \tiny(91.7)  \\
ELECTRA$_\text{base}$ & 68.4 \tiny(68.0) & 84.8 \tiny(84.6) & 73.1 \tiny(72.7) & 87.1 \tiny(86.9) & 22.6 \tiny(21.7) & 45.0 \tiny(43.8) & 87.5 \tiny(87.0) & 92.5 \tiny(92.3) & 86.9 \tiny(86.6) & 91.8 \tiny(91.7) \\
MacBERT$_\text{base}$ & 68.5 \tiny(67.3) & 87.9 \tiny(87.1) & 73.2 \tiny(72.4) & 89.5 \tiny(89.2) & {\bf 30.2} \tiny(26.4) & 54.0 \tiny(52.2) & 89.4 \tiny(89.2) & 94.3 \tiny(94.1) & 89.5 \tiny(88.7) & 93.8 \tiny(93.5) \\
PERT$_\text{base}$ & 68.5 \tiny(68.1) & 87.2 \tiny(87.1) & 72.8 \tiny(72.5) & 89.2 \tiny(89.0) & 28.7 \bf\tiny(28.2) & 55.4 \tiny(53.7) & {89.5} \tiny(88.9) & 93.9 \tiny(93.6) & 89.0 (88.5) & 93.5 \tiny(93.2) \\
\bf LERT$_\text{base}$ 
& \bf 69.2 \tiny(68.4) & \bf 88.1 (87.9) & \bf 73.5 \tiny(72.8) & \bf 89.7 \tiny(89.4) & 27.7 \tiny(26.7) & \bf 55.9 \tiny(54.6) 
& \bf 90.5 \tiny(90.2) & \bf 95.1 \tiny(94.9) & \bf 90.5 \tiny(90.2) & \bf 94.9 \tiny(94.7) \\
\midrule 
RoBERTa$_\text{large}$ & 68.5 \tiny(67.6) & 88.4 \tiny(87.9) & 74.2 \tiny(72.4) & 90.6 \tiny(90.0) & 31.5 \tiny(30.1) & 60.1 \tiny(57.5)  & 89.6 \tiny(89.1) & 94.8 \tiny(94.4) & 89.6 \tiny(88.9) & 94.5 \tiny(94.1) \\
ELECTRA$_\text{large}$ & 69.1 \tiny(68.2) & 85.2 \tiny(84.5) & 73.9 \tiny(72.8) & 87.1 \tiny(86.6) & 23.0 \tiny(21.6) & 44.2 \tiny(43.2) & 88.8 \tiny(88.7) & 93.3 \tiny(93.2) & 88.8 \tiny(88.2) & 93.6 \tiny(93.2) \\
MacBERT$_\text{large}$ 		& 70.7 \tiny(68.6) & 88.9 \tiny(88.2) & 74.8 \tiny(73.2) & 90.7 \tiny(90.1) & 31.9 \tiny(29.6) & 60.2 \tiny(57.6) & 91.2 \tiny(90.8) & 95.6 \tiny(95.3) & {\bf 91.7} \tiny(90.9) & 95.6 \tiny(95.3) \\
PERT$_\text{large}$ & \bf 72.2 \tiny(71.0) & 89.4 \tiny(88.8) & \bf 76.8 \tiny(75.5) & 90.7 \tiny(90.4) & \bf 32.3 \tiny(30.9) & 59.2 \tiny(58.1) & 90.9 \tiny(90.8) & 95.5 \tiny(95.2) & 91.1 \tiny(90.7) & 95.2 \tiny(95.1) \\
\bf LERT$_\text{large}$ 
& 71.2 \tiny(70.5) & \bf 89.5 \tiny(89.1) & 75.6 \tiny(75.1) & \bf 90.9 \tiny(90.6) & 32.3 \tiny(29.7) & \bf 61.2 \tiny(59.2) 
& \bf 91.6 \tiny(91.3) & \bf 96.1 \tiny(95.8) & 91.5 \bf\tiny(91.1) & \bf95.9 \tiny(95.5) \\
\bottomrule
\end{tabular}
\end{center}
\caption{\label{results-mrc} Experimental results on MRC tasks: CMRC 2018 (Simplified Chinese) and DRCD (Traditional Chinese). We report both the maximum and average scores (in parenthesis) for each set. Overall best performances are depicted in boldface (base-level and large-level are marked individually). }
\end{table*}

\begin{table*}[ht]
\tiny
\begin{center}
\begin{tabular}{l cc cc cc cc cc }
\toprule
\multirow{2}*{\bf System} & \multicolumn{2}{c}{\centering \bf XNLI} & \multicolumn{2}{c}{\centering \bf LCQMC} & \multicolumn{2}{c}{\centering \bf BQ Corpus} & \multicolumn{2}{c}{\centering \bf ChnSentiCorp}  & \bf TNEWS  & \bf OCNLI \\
 & Dev & Test  & Dev & Test & Dev & Test & Dev & Test & Dev & Dev \\
\midrule
BERT$_\text{base}$ & 79.4 \tiny(78.6) & 78.7 \tiny(78.3)  & 89.6 \tiny(89.2) & 87.1 \tiny(86.6) & {\bf 86.4} \tiny(85.5)  & {\bf 85.3} \tiny(84.8) & {\bf 95.4} \tiny(94.6) & 95.3 \tiny(94.8)  & 57.0 \tiny(56.6) & 76.0 \tiny(75.3) \\
RoBERTa$_\text{base}$ & 80.0 \tiny(79.2) & 78.8 \tiny(78.3)  & 89.0 \tiny(88.7) & 86.4 \tiny(86.1) & 86.0 \tiny(85.4) & 85.0 \tiny(84.6) & 94.9 \tiny(94.6) & 95.6 \tiny(94.9) & 57.4 \tiny(56.9)  & 76.5 \tiny(76.0)  \\
ELECTRA$_\text{base}$ 	& 77.9 \tiny(77.0) & 78.4 \tiny(77.8) & \bf 90.2 \tiny(89.8) & \bf 87.6 \tiny(87.3) & 84.8 \tiny(84.7) & 84.5 \tiny(84.0) & 93.8 \tiny(93.0) & 94.5 \tiny(93.5)  & 56.1 \tiny(55.7) & 76.1 \tiny(75.8) \\  
MacBERT$_\text{base}$ & \bf 80.3 \tiny(79.7) & 79.3 \tiny(78.8) & 89.5 \tiny(89.3) & 87.0 \tiny(86.5) & 86.0 \tiny(85.5) & 85.2 \bf\tiny(84.9) & 95.2 \bf\tiny(94.8) & 95.6 \tiny(94.9)  & 57.4 
\bf\tiny(57.1)  & 77.0 \tiny(76.5)  \\
PERT$_\text{base}$ & 78.8 \tiny(78.1) & 78.1 \tiny(77.7) & 88.8 \tiny(88.3) & 86.3 \tiny(86.0) & 84.9 \tiny(84.8) & 84.3 \tiny(84.1) & 94.0 \tiny(93.7) & 94.8 \tiny(94.1) & 56.7 \tiny(56.1) & 75.3 \tiny(74.8) \\
\bf LERT$_\text{base}$ 
& 80.2 \tiny(79.5) & \bf 79.8 \tiny(79.3) 
& 89.5 \tiny(89.2) & 86.6 \tiny(86.4) 
& 85.9 \bf\tiny(85.6) & 85.1 \bf \tiny(84.9) 
& 94.9 \tiny(94.7) & \bf 95.9 \tiny(95.2) 
& \bf 57.5 \tiny(57.1) & \bf 78.2 \tiny(77.5) \\
\midrule
RoBERTa$_\text{large}$ & 82.1 \tiny(81.3) & 81.2 \tiny(80.6)  & 90.4 \tiny(90.0) & 87.0 \tiny(86.8) & 86.3 \tiny(85.7) & {\bf 85.8} \tiny(84.9) & {\bf 95.8} \tiny(94.9) & 95.8 \tiny(94.9)  & 58.8 \tiny(58.4)  & 78.5 \tiny(78.2)  \\
ELECTRA$_\text{large}$ 	& 81.5 \tiny(80.8) & 81.0 \bf\tiny(80.9) & \bf 90.7 \tiny(90.4) & 87.3 \bf\tiny(87.2) & \bf 86.7 \tiny(86.2) & 85.1 \tiny(84.8) & 95.2 \tiny(94.6) & 95.3 \tiny(94.8)   & 57.2 \tiny(56.9) & 78.8 \tiny(78.4)  \\
MacBERT$_\text{large}$ & \bf 82.4 \tiny(81.8) & {\bf 81.3} \tiny(80.6) & 90.6 \tiny(90.3) & {\bf 87.6} \tiny(87.1) & 86.2 \tiny(85.7) & 85.6 \bf \tiny(85.0) & 95.7 \bf \tiny(95.0) & 95.9 \tiny(95.1)  & \bf 59.0 \tiny(58.8) & 79.0 \tiny(78.7) \\
PERT$_\text{large}$ & 81.0 \tiny(80.4) & 80.4 \tiny(80.1) & 90.0 \tiny(89.7) & 87.2 \tiny(86.9) & 86.3 \tiny(85.8) & 85.0 \tiny(84.8) & 94.5 \tiny(94.0) & 95.3 \tiny(94.8) & 57.4 \tiny(57.2) & 78.1 \tiny(77.8) \\
\bf LERT$_\text{large}$ 
& 81.7 \tiny(81.2) & 81.0 \tiny(80.7)
& 90.2 \tiny(90.0) & 87.3 \tiny(86.9)
& 86.6 \tiny(86.0) & 85.1 \tiny(84.7)
& 95.6 \tiny(94.9) & \bf 96.2 \tiny(95.4) 
& 58.7 \tiny(58.5) & \bf 79.4 \tiny(78.9) \\
\bottomrule
\end{tabular}
\end{center}
\caption{\label{results-tc} Experimental results on text classification tasks (including natural language inference tasks): XNLI, LCQMC, BQ Corpus, ChnSentiCorp, TNEWS, and OCNLI. }
\end{table*}

\begin{table}[h]
\small
\begin{center}
\begin{tabular}{p{2cm} p{2cm}<\centering p{2cm}<\centering}
\toprule
\multirow{1}*{\bf System} & \multicolumn{1}{c}{\centering \bf MSRA-NER} & \multicolumn{1}{c}{\centering \bf People's Daily}  \\
\midrule
BERT$_\text{base}$ 		& 95.3 \tiny(94.9) & 95.3 \tiny (95.1) \\
RoBERTa$_\text{base}$ 	& 95.5 \tiny(95.1) & 95.1 \tiny(94.9) \\
ELECTRA$_\text{base}$ 	& 95.4 \tiny(95.0) & 95.1 \tiny(94.9) \\  
MacBERT$_\text{base}$ 	& 95.3 \tiny(95.1) & 95.2 \tiny(94.9) \\
PERT$_\text{base}$  		& 95.6 \tiny(95.3) & 95.3 \tiny(95.1) \\
\bf LERT$_\text{base}$  & \bf 95.7 \tiny(95.4) & \bf 95.6 \tiny(95.4) \\
\midrule
RoBERTa$_\text{large}$ 	& 95.5 \tiny(95.5) & 95.7 \tiny(95.4) \\
ELECTRA$_\text{large}$ 	& 95.0 \tiny(94.8) & 94.9 \tiny(94.8) \\  
MacBERT$_\text{large}$ 	& 96.2 \tiny(95.9) & 95.8 \tiny(95.7) \\
PERT$_\text{large}$ 		& 96.2 \bf \tiny(96.0) & 96.1 \tiny(95.8) \\
\bf LERT$_\text{large}$ & \bf 96.3 \tiny(96.0) & \bf 96.3 \tiny(96.0) \\
\bottomrule
\end{tabular}
\end{center}
\caption{\label{results-ner} Experimental results (F-score) on NER tasks.}
\end{table}

We use a universal initial learning rate for each task for the same model size, with 5e-5 for small-sized models, 3e-5 for base-sized models, and 2e-5 for large-sized models.
Other details for task fine-tuning are shown in Table \ref{ft-hyper}.
The implementations are based on original BERT.\footnote{\url{https://github.com/google-research/bert}}

%%%%%%%%%%%%%%
\subsection{Main Results}
We mainly compare our results with pre-trained language models that use a similar amount of training data.
Experimental results on base-sized and large-sized models are shown in Table \ref{results-mrc}, \ref{results-tc}, and \ref{results-ner}.

For machine reading comprehension tasks, LERT yields significant improvements over various pre-trained language models by a large margin on both base-sized and large-sized LERT.
This indicates that LERT can better handle complex task that requires various types of linguistic knowledge (e.g., machine reading comprehension). 

For text classification tasks, the results are varied.
We can see that LERT yields the best performance on several tasks, such as TNEWS, OCNLI, ChnSentiCorp, etc.
For other tasks, the average results are competitive against the best-performing pre-trained language model.
Unlike the MRC task, the text classification tasks are usually determined by very few words in the input sequence, such as sentiment words, negation words, etc.
In this context, we speculate that the additional linguistic knowledge introduced in LERT is not that useful for further improving classification accuracy.
Nonetheless, we can see that LERT still provides decent scores on several classification tasks.

\begin{table*}[htbp]
\begin{center}
\small
\begin{tabular}{l c | cc cc | cccc cc | cc }
\toprule
\multirow{2}*{\bf System} & \multirow{2}*{\bf Param} & \multicolumn{2}{c}{\centering \bf CMRC 2018} & \multicolumn{2}{c}{\centering \bf DRCD} & {\bf XNLI} & {\bf LC} & {\bf BQ} &  {\bf CSC} &  {\bf TN} &  {\bf OC} &  {\bf MSRA} &  {\bf PD}  \\
& & EM & F1 & EM & F1 & ACC & ACC  & ACC & ACC & ACC & ACC & F & F \\
\midrule
RBT3 			& 38M & 62.2 & 81.8 & 75.0 & 83.9 & 72.3 & 85.1 & 83.3 & 92.8 & - & - & - & - \\
ELT$_\text{small+}$ & 12M 	& \bf 68.5 & \bf 85.2 & 82.9 & 88.7 & 74.6 & 85.8 & 82.1 & 93.6 & - & - & - & - \\
\midrule
ELT$_\text{small}$ 	& 12M 	& \em 67.8 & 83.4 & 79.0 & 85.8 & 73.1 & \bf \em 85.9 & 82.0 & \bf \em 94.3 & - & - & - & - \\
BERT$_\text{small}$ 	& 15M 	& 65.3 & 83.9 & 81.7 & 88.1 & 74.6 & 85.7 & 83.0 & 93.6 & \bf \em 55.2 & 70.8 & 91.8 & 91.5 \\ 
LERT$_\text{small}$ 	& 15M 	& \em 67.8 & \bf \em 85.2 & \bf \em 83.2 & \bf \em 89.4 
& \bf \em 75.2 & 85.3 & \bf \em 83.4 & 94.0 
& 54.9 & \bf \em 71.0 & \bf \em 92.3 & \bf \em 92.1 \\
\bottomrule
\end{tabular}
\caption{\label{results-small-models} Test results on small models. ELT: ELECTRA. Overall best scores are depicted in boldface, and the comparable best score (the same training data size) is shown in italics.} 
\end{center}
\end{table*}

\begin{table*}[htbp]
\begin{center}
\small
\begin{tabular}{l  cc cc  cccc  cc  cc  c}
\toprule
\multirow{2}*{\bf System} & \multicolumn{2}{c}{\centering \bf CMRC} & \multicolumn{2}{c}{\centering \bf DRCD} & {\bf XNLI} & {\bf LC} & {\bf BQ} &  {\bf CSC} &  {\bf TN} &  {\bf OC} &  {\bf MSRA} &  {\bf PD} & \multirow{2}*{\bf Average}  \\
& EM & F1 & EM & F1 & ACC & ACC  & ACC & ACC & ACC & ACC & F & F \\
\midrule
Baseline		& 66.8 & 86.7 & 89.0 & 94.1 & 78.1 & 88.7 & 85.1 & 94.2 & 56.4 & 76.0 & 94.6 & 94.4 & 83.58 \\
\quad + POS		& 67.0 & 86.9 & 89.0 & 93.9 & 78.4 & 88.9 & 85.1 & 94.2 & 56.3 & 75.6 & 95.1 & 95.2 & 83.72 \tiny(+0.14) \\
\quad + NER		& 67.0 & 87.2 & 89.4 & 94.3 & 78.5 & 89.1 & 85.3 & 94.2 & 56.8 & 76.0 & 95.5 & 95.4 & 83.97 \tiny(+0.39)\\
\quad + DEP		& 67.2 & 86.9 & 89.3 & 94.2 & 78.6 & 88.9 & 85.0 & 94.0 & 56.0 & 76.0 & 94.9 & 95.0 & 83.72 \tiny(+0.14) \\
\quad + All		& 67.5 & 87.3 & 89.5 & 94.4 & 78.8 & 88.9 & 85.2 & 94.2 & 56.7 & 76.4 & 95.4 & 95.4 & 84.03 \tiny(+0.45) \\
LERT$_\text{base}$	& 68.4 & 87.9 & 90.2 & 94.9 & 79.5 & 89.2 & 85.6 & 94.7 & 57.1 & 77.5 & 95.4 & 95.4 & 84.65 \tiny(+1.07) \\
\bottomrule
\end{tabular}
\caption{\label{results-ablation} Ablation results on using different linguistic features. We report five-run average scores on the development set of each task. The reported results are on {\em base}-sized PLMs, trained with 500k steps (except for LERT$_\text{base}$).}
\end{center}
\end{table*}

For named entity recognition tasks, we can see that LERT yields the best performance on both tasks, including base-sized and large-sized variants.
The results are expected because the NER task is added into the pre-training stage as one of the linguistic tasks.
Further analyses are presented in Section \ref{sec-analysis}.

Overall, LERT yields significant improvements on MRC and NER tasks and achieves competitive performance on TC tasks over various pre-trained language models.

%%%%%%%%%%%%%%
\subsection{Results on Small Models}

Along with conventional base-sized and large-sized LERT, we also train a small-sized LERT$_\text{small}$.
Unlike base-sized and large-sized models, small models are usually not comparable to the previous work due to various types of model structure, including hidden size, number of layers, number of attention heads, etc.
In this paper, to make the results comparable, we also train a BERT$_\text{small}$, which shares the same training recipe with LERT$_\text{small}$, except that it is only trained on the MLM task.
The experimental results are shown in Table \ref{results-small-models}.

Similar to the base-sized and large-sized models, LERT yields consistent improvements on MRC and NER tasks and moderate improvements on classification tasks, which further demonstrates that linguistic knowledge preference for different tasks differs.
We also compare LERT$_\text{small}$ with RBT3 (38M parameters), ELECTRA$_\text{small}$ (12M), and ELECTRA$_\text{small+}$ (12M).
Note that ELECTRA \citep{clark2020electra} uses embedding decomposition that projects word embedding into a smaller vector and then uses a fully-connected layer to project into hidden size.
However, LERT does not apply this approach, resulting in a little bit more parameter sizes than ELECTRA.
The results show that LERT$_\text{small}$ performs better on a majority of downstream tasks over ELECTRA$_\text{small}$ and RBT3, and even better than ELECTRA$_\text{small+}$, which was trained on 180G pre-training data.

%%%%%%%%%%%%%%%%%%%%%%%%%%%%%%%%%%%%%%%%
\section{Analysis}\label{sec-analysis}

%%%%%%%%%%%%%%
\subsection{Ablation Study}
In order to identify the effectiveness of each linguistic task, in this section, we perform the ablation study on LERT.
We add each linguistic task on top of MLM (baseline) to verify their effectiveness individually.\footnote{We do not perform task warmup for these experiments to keep comparisons as pure as possible.}
The results are shown in Table \ref{results-ablation}.
As we can see that all three types of linguistic features contribute to the overall improvement positively, where the NER features are the most important, especially for downstream NER tasks.
Furthermore, using all three linguistic features yields another boost in the final performance, where all downstream tasks yield consistent improvements.

%%%%%%%%%%%%%%
\subsection{Effect of Linguistic Task Order}\label{sec-task-order}
\begin{figure}[htp]
  \centering
  \includegraphics[width=1\columnwidth]{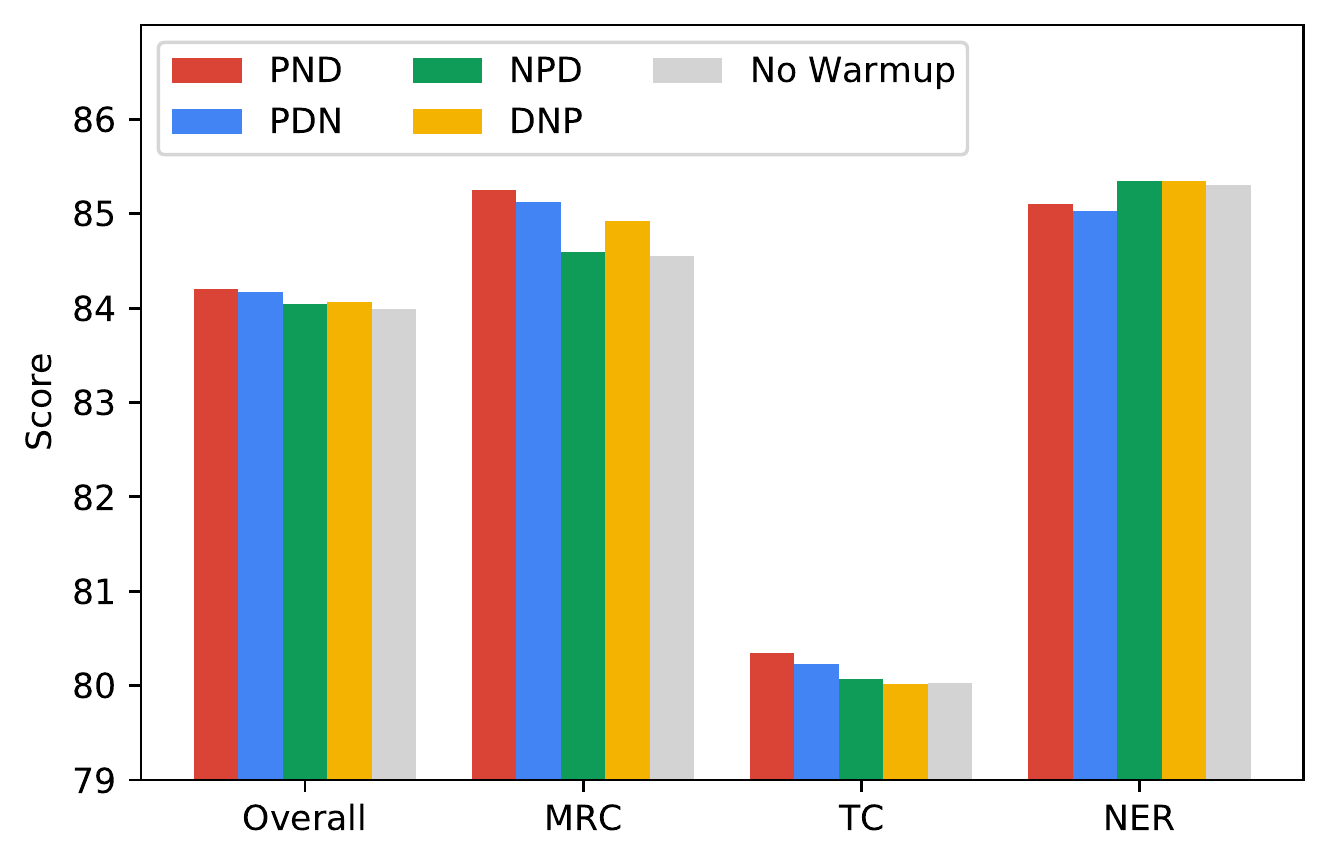}
  \caption{\label{fig-task-order} Effect of different linguistic task order in LIP. Note that the scores of NER are subtracted by 10 for clarity. All models are trained with 1M steps.} 
\end{figure}

For better pre-training LERT, we propose a linguistically-informed pre-training strategy, which learns basic linguistic knowledge faster (POS) than the high-level knowledge (NER and DEP), forming a ``PND'' scheme.\footnote{We use the initials of each linguistic task to denote a specific warmup scheme. Concretely, ``PND'' means the POS features learn fastest, then NER, followed by DEP.}
To further demonstrate its effectiveness, we also trained another three LERT$_\text{base}$ models that use different task warmup strategies, including PDN, NPD, and DNP, which indicates the different warmup order of linguistic tasks.
To demonstrate the effectiveness of the task warmup strategy, we add another ``no warmup'' experiment that uses Equation \ref{eq-overall-loss-vanilla} for training.
The results are shown in Figure \ref{fig-task-order}.
Overall, the original implementation in LERT (i.e., ``PND'' scheme) yields the best performance among all variants, where we conclude our findings as follows.
\begin{itemize}
	\item By comparing ``PND'', ``PDN'', and ``NPD'', we discover that results are better when ``POS'' feature learns faster, which matches our intuitive thinking that fundamental knowledge should be learned faster.
	\item ``NPD'' scheme yields the best performance on NER tasks, suggesting that if the pre-trained task is directly associated with the downstream tasks, it is always better to learn from the beginning.
	\item ``PND'' and ``PDN'' yield better overall performance than the others, indicating that faster basic knowledge (POS) learning is helpful for better high-level knowledge learning. 
	\item All task warmup schemes show better overall performance than no task warmup strategy (i.e., equal weights for each task).
\end{itemize}

%%%%%%%%%%%%%%
\subsection{Effect of Linguistical Masking}
In this paper, we use linguistical multi-task learning to formalize LERT, where the linguistic knowledge is used at the output as labels in the pre-training stage.
However, we wonder if linguistic knowledge can be applied at the input as a hint of masking and whether it is more effective than LERT's implementation.

To achieve this goal, we extend the original masked language model as a linguistic masked language model (LMLM).
In traditional MLM, the masking token is {\tt [MASK]}, which does not carry any linguistic information.
In LMLM, we further incorporate linguistic tags into {\tt [MASK]}, forming a list of different masking tokens.
For example, if the POS tag of the masked token is a noun, then the corresponding masked token is set as {\tt [MASK-POS-n]}.
In this way, the model is informed with additional linguistic hints for the masked tokens.
We train three types of LMLM w.r.t. each type of linguistic knowledge.
Also, we set up two additional settings, called ``All'' (incorporating all three types of linguistic tags into {\tt [MASK]}) and ``Mix'' (randomly assign one linguistic tag into {\tt [MASK]}.
We use the same training settings as in Table \ref{results-ablation}.
The results of LMLM are listed in Table \ref{results-linguistic-masking}.

As we can see that incorporating NER tags into the masked token yields improvement over vanilla MLM, especially for the NER tasks.
However, for most of the other settings, LMLM does not yield consistent improvements.
By comparing to the results in Table \ref{results-ablation}, we can see that exploiting linguistic knowledge as the training target yields consistent and significant improvements over vanilla MLM and LMLMs.
For example, using NER tags as the training target (average score: 83.97, in Table \ref{results-ablation}) yields better performance than in LMLM (average score: 83.70).
These results indicate that the design of LERT is valid.
\begin{table}[h]
\small
\begin{center}
\begin{tabular}{l  ccccc}
\toprule
\multirow{1}*{\bf System} & \multicolumn{1}{c}{\centering \bf MRC} & \multicolumn{1}{c}{\centering \bf TC} & \multicolumn{1}{c}{\centering \bf NER} & \bf Average \\
\midrule
MLM 				& 84.2 & 79.8 & 94.5 & 83.58 \\
\quad + POS-mask  	& 83.1 & 79.4 & 94.6 & 83.14 \\
\quad + NER-mask  	& 83.9 & 79.9 & 94.8 & 83.70 \\
\quad + DEP-mask  	& 83.7 & 79.7 & 94.5 & 83.43 \\
\quad + All-mask  	& 83.5 & 79.5 & 94.5 & 83.26 \\
\quad + Mix-mask  	& 83.9 & 79.5 & 94.4 & 83.40 \\
\bottomrule
\end{tabular}
\end{center}
\caption{\label{results-linguistic-masking} Results of linguistical masking (LMLM).}
\end{table}

%%%%%%%%%%%%%%%%%%%%%%%%%%%%%%%%%%%%%%%%
\section{Conclusion}
In this paper, we propose a new pre-trained language model called LERT, which directly incorporates three types of linguistic features and performs multi-task pre-training along with the masked language model.
Three types of common linguistic features, including POS, NER, and DEP, are generated by LTP, and LERT learns to predict both the original word and its linguistic tags for masked tokens.
To better acquire linguistic knowledge, we also propose a linguistically-informed pre-training strategy that learns basic linguistic faster than the high-level ones, which we empirically find useful.
We carried out extensive and robust experiments on ten Chinese natural language understanding tasks.
The experimental results show that LERT could bring significant improvements over various comparable pre-trained language models, demonstrating that linguistic knowledge can still boost the performance of pre-trained language models, especially for small-sized models.

In the future, we are going to incorporate more types of linguistic features into pre-trained language models, such as semantic dependency parsing, etc. 
Also, as the proposed task warmup strategy seems to be generally useful, we are going to investigate if it is helpful to other multi-task learning scenarios.

%%%%%%%%%%%%%%%%%%%%%%%%%%%%%%%%%%%%%%%%%
\section*{Acknowledgments}\label{ack}
% We would like to thank all anonymous reviewers for their constructive comments. 
Yiming Cui would like to thank continuous support from Google's TPU Research Cloud (TRC) program for Cloud TPU access.

\bibliography{custom}

\begin{thebibliography}{35}
\expandafter\ifx\csname natexlab\endcsname\relax\def\natexlab#1{#1}\fi

\bibitem[{Che et~al.(2010)Che, Li, and Liu}]{che2010ltp}
Wanxiang Che, Zhenghua Li, and Ting Liu. 2010.
\newblock Ltp: A chinese language technology platform.
\newblock In \emph{Proceedings of the 23rd International Conference on
  Computational Linguistics: Demonstrations}, pages 13--16. Association for
  Computational Linguistics.

\bibitem[{Chen et~al.(2018)Chen, Chen, Liu, Yang, Lu, and
  Tang}]{chen-etal-2018-bq}
Jing Chen, Qingcai Chen, Xin Liu, Haijun Yang, Daohe Lu, and Buzhou Tang. 2018.
\newblock \href {https://www.aclweb.org/anthology/D18-1536} {The {BQ} corpus: A
  large-scale domain-specific {C}hinese corpus for sentence semantic
  equivalence identification}.
\newblock In \emph{Proceedings of the 2018 Conference on Empirical Methods in
  Natural Language Processing}, pages 4946--4951, Brussels, Belgium.
  Association for Computational Linguistics.

\bibitem[{Clark et~al.(2020)Clark, Luong, Le, and Manning}]{clark2020electra}
Kevin Clark, Minh-Thang Luong, Quoc~V. Le, and Christopher~D. Manning. 2020.
\newblock \href {https://openreview.net/pdf?id=r1xMH1BtvB} {{ELECTRA}:
  Pre-training text encoders as discriminators rather than generators}.
\newblock In \emph{ICLR}.

\bibitem[{Conneau et~al.(2018)Conneau, Rinott, Lample, Williams, Bowman,
  Schwenk, and Stoyanov}]{conneau2018xnli}
Alexis Conneau, Ruty Rinott, Guillaume Lample, Adina Williams, Samuel~R.
  Bowman, Holger Schwenk, and Veselin Stoyanov. 2018.
\newblock Xnli: Evaluating cross-lingual sentence representations.
\newblock In \emph{Proceedings of the 2018 Conference on Empirical Methods in
  Natural Language Processing}. Association for Computational Linguistics.

\bibitem[{Cui et~al.(2021)Cui, Che, Liu, Qin, and Yang}]{chinese-bert-wwm}
Yiming Cui, Wanxiang Che, Ting Liu, Bing Qin, and Ziqing Yang. 2021.
\newblock \href {https://doi.org/10.1109/TASLP.2021.3124365} {Pre-training with
  whole word masking for chinese bert}.
\newblock \emph{IEEE/ACM Transactions on Audio, Speech, and Language
  Processing}, 29:3504--3514.

\bibitem[{Cui et~al.(2019)Cui, Liu, Che, Xiao, Chen, Ma, Wang, and
  Hu}]{cui-emnlp2019-cmrc2018}
Yiming Cui, Ting Liu, Wanxiang Che, Li~Xiao, Zhipeng Chen, Wentao Ma, Shijin
  Wang, and Guoping Hu. 2019.
\newblock \href {https://www.aclweb.org/anthology/D19-1600} {A
  {S}pan-{E}xtraction {D}ataset for {C}hinese {M}achine {R}eading
  {C}omprehension}.
\newblock In \emph{Proceedings of the 2019 Conference on Empirical Methods in
  Natural Language Processing and the 9th International Joint Conference on
  Natural Language Processing (EMNLP-IJCNLP)}, pages 5886--5891, Hong Kong,
  China. Association for Computational Linguistics.

\bibitem[{Cui et~al.(2022)Cui, Yang, and Liu}]{cui2022pert}
Yiming Cui, Ziqing Yang, and Ting Liu. 2022.
\newblock \href {http://arxiv.org/abs/2203.06906} {Pert: Pre-training bert with
  permuted language model}.
\newblock \emph{arXiv preprint arXiv:2203.06906}.

\bibitem[{Devlin et~al.(2019)Devlin, Chang, Lee, and
  Toutanova}]{devlin-etal-2019-bert}
Jacob Devlin, Ming-Wei Chang, Kenton Lee, and Kristina Toutanova. 2019.
\newblock \href {https://www.aclweb.org/anthology/N19-1423} {{BERT}:
  Pre-training of deep bidirectional transformers for language understanding}.
\newblock In \emph{Proceedings of the 2019 Conference of the North {A}merican
  Chapter of the Association for Computational Linguistics: Human Language
  Technologies, Volume 1 (Long and Short Papers)}, pages 4171--4186,
  Minneapolis, Minnesota. Association for Computational Linguistics.

\bibitem[{He et~al.(2021)He, Liu, Gao, and Chen}]{he2021deberta}
Pengcheng He, Xiaodong Liu, Jianfeng Gao, and Weizhu Chen. 2021.
\newblock \href {https://openreview.net/forum?id=XPZIaotutsD} {{DeBERTa}:
  Decoding-enhanced bert with disentangled attention}.
\newblock In \emph{International Conference on Learning Representations}.

\bibitem[{Hewitt and Manning(2019)}]{hewitt-manning-2019-structural}
John Hewitt and Christopher~D. Manning. 2019.
\newblock \href {https://doi.org/10.18653/v1/N19-1419} {{A} structural probe
  for finding syntax in word representations}.
\newblock In \emph{Proceedings of the 2019 Conference of the North {A}merican
  Chapter of the Association for Computational Linguistics: Human Language
  Technologies, Volume 1 (Long and Short Papers)}, pages 4129--4138,
  Minneapolis, Minnesota. Association for Computational Linguistics.

\bibitem[{Hu et~al.(2020)Hu, Richardson, Xu, Li, Kuebler, and Moss}]{ocnli}
Hai Hu, Kyle Richardson, Liang Xu, Lu~Li, Sandra Kuebler, and Larry Moss. 2020.
\newblock \href {https://arxiv.org/abs/2010.05444} {Ocnli: Original chinese
  natural language inference}.
\newblock In \emph{Findings of EMNLP}.

\bibitem[{Jawahar et~al.(2019)Jawahar, Sagot, and
  Seddah}]{jawahar-etal-2019-bert}
Ganesh Jawahar, Beno{\^\i}t Sagot, and Djam{\'e} Seddah. 2019.
\newblock \href {https://doi.org/10.18653/v1/P19-1356} {What does {BERT} learn
  about the structure of language?}
\newblock In \emph{Proceedings of the 57th Annual Meeting of the Association
  for Computational Linguistics}, pages 3651--3657, Florence, Italy.
  Association for Computational Linguistics.

\bibitem[{Kingma and Ba(2014)}]{kingma2014adam}
Diederik Kingma and Jimmy Ba. 2014.
\newblock Adam: A method for stochastic optimization.
\newblock \emph{arXiv preprint arXiv:1412.6980}.

\bibitem[{Kovaleva et~al.(2019)Kovaleva, Romanov, Rogers, and
  Rumshisky}]{kovaleva-etal-2019-revealing}
Olga Kovaleva, Alexey Romanov, Anna Rogers, and Anna Rumshisky. 2019.
\newblock \href {https://doi.org/10.18653/v1/D19-1445} {Revealing the dark
  secrets of {BERT}}.
\newblock In \emph{Proceedings of the 2019 Conference on Empirical Methods in
  Natural Language Processing and the 9th International Joint Conference on
  Natural Language Processing (EMNLP-IJCNLP)}, pages 4365--4374, Hong Kong,
  China. Association for Computational Linguistics.

\bibitem[{Lan et~al.(2020)Lan, Chen, Goodman, Gimpel, Sharma, and
  Soricut}]{lan2019albert}
Zhenzhong Lan, Mingda Chen, Sebastian Goodman, Kevin Gimpel, Piyush Sharma, and
  Radu Soricut. 2020.
\newblock \href {https://openreview.net/forum?id=H1eA7AEtvS} {Albert: A lite
  bert for self-supervised learning of language representations}.
\newblock In \emph{International Conference on Learning Representations}, pages
  1--17.

\bibitem[{Levow(2006)}]{levow-2006-third}
Gina-Anne Levow. 2006.
\newblock \href {https://www.aclweb.org/anthology/W06-0115} {The third
  international {C}hinese language processing bakeoff: Word segmentation and
  named entity recognition}.
\newblock In \emph{Proceedings of the Fifth {SIGHAN} Workshop on {C}hinese
  Language Processing}, pages 108--117, Sydney, Australia. Association for
  Computational Linguistics.

\bibitem[{Liu et~al.(2019{\natexlab{a}})Liu, Gardner, Belinkov, Peters, and
  Smith}]{liu-etal-2019-linguistic}
Nelson~F. Liu, Matt Gardner, Yonatan Belinkov, Matthew~E. Peters, and Noah~A.
  Smith. 2019{\natexlab{a}}.
\newblock \href {https://doi.org/10.18653/v1/N19-1112} {Linguistic knowledge
  and transferability of contextual representations}.
\newblock In \emph{Proceedings of the 2019 Conference of the North {A}merican
  Chapter of the Association for Computational Linguistics: Human Language
  Technologies, Volume 1 (Long and Short Papers)}, pages 1073--1094,
  Minneapolis, Minnesota. Association for Computational Linguistics.

\bibitem[{Liu et~al.(2021)Liu, Fu, Zhang, and Xiao}]{liu-etal-2021-lexicon}
Wei Liu, Xiyan Fu, Yue Zhang, and Wenming Xiao. 2021.
\newblock \href {https://doi.org/10.18653/v1/2021.acl-long.454} {Lexicon
  enhanced {C}hinese sequence labeling using {BERT} adapter}.
\newblock In \emph{Proceedings of the 59th Annual Meeting of the Association
  for Computational Linguistics and the 11th International Joint Conference on
  Natural Language Processing (Volume 1: Long Papers)}, pages 5847--5858,
  Online. Association for Computational Linguistics.

\bibitem[{Liu et~al.(2018)Liu, Chen, Deng, Zeng, Chen, Li, and
  Tang}]{liu2018lcqmc}
Xin Liu, Qingcai Chen, Chong Deng, Huajun Zeng, Jing Chen, Dongfang Li, and
  Buzhou Tang. 2018.
\newblock Lcqmc: A large-scale chinese question matching corpus.
\newblock In \emph{Proceedings of the 27th International Conference on
  Computational Linguistics}, pages 1952--1962.

\bibitem[{Liu et~al.(2019{\natexlab{b}})Liu, Ott, Goyal, Du, Joshi, Chen, Levy,
  Lewis, Zettlemoyer, and Stoyanov}]{liu2019roberta}
Yinhan Liu, Myle Ott, Naman Goyal, Jingfei Du, Mandar Joshi, Danqi Chen, Omer
  Levy, Mike Lewis, Luke Zettlemoyer, and Veselin Stoyanov. 2019{\natexlab{b}}.
\newblock Roberta: A robustly optimized bert pretraining approach.
\newblock \emph{arXiv preprint arXiv:1907.11692}.

\bibitem[{Manning et~al.(2014)Manning, Surdeanu, Bauer, Finkel, Bethard, and
  McClosky}]{manning-etal-2014-stanford}
Christopher Manning, Mihai Surdeanu, John Bauer, Jenny Finkel, Steven Bethard,
  and David McClosky. 2014.
\newblock \href {https://doi.org/10.3115/v1/P14-5010} {The {S}tanford
  {C}ore{NLP} natural language processing toolkit}.
\newblock In \emph{Proceedings of 52nd Annual Meeting of the Association for
  Computational Linguistics: System Demonstrations}, pages 55--60, Baltimore,
  Maryland. Association for Computational Linguistics.

\bibitem[{Mikolov et~al.(2013)Mikolov, Sutskever, Chen, Corrado, and
  Dean}]{mikolov-etal-2013}
Tomas Mikolov, Ilya Sutskever, Kai Chen, Greg~S Corrado, and Jeff Dean. 2013.
\newblock \href
  {http://papers.nips.cc/paper/5021-distributed-representations-of-words-and-phrases-and-their-compositionality.pdf}
  {Distributed representations of words and phrases and their
  compositionality}.
\newblock In C.~J.~C. Burges, L.~Bottou, M.~Welling, Z.~Ghahramani, and K.~Q.
  Weinberger, editors, \emph{Advances in Neural Information Processing Systems
  26}, pages 3111--3119. Curran Associates, Inc.

\bibitem[{Pennington et~al.(2014)Pennington, Socher, and
  Manning}]{pennington-etal-2014-glove}
Jeffrey Pennington, Richard Socher, and Christopher Manning. 2014.
\newblock \href {https://doi.org/10.3115/v1/D14-1162} {{G}lo{V}e: Global
  vectors for word representation}.
\newblock In \emph{Proceedings of the 2014 Conference on Empirical Methods in
  Natural Language Processing ({EMNLP})}, pages 1532--1543, Doha, Qatar.
  Association for Computational Linguistics.

\bibitem[{Peters et~al.(2018)Peters, Neumann, Iyyer, Gardner, Clark, Lee, and
  Zettlemoyer}]{peters-etal-2018-deep}
Matthew~E. Peters, Mark Neumann, Mohit Iyyer, Matt Gardner, Christopher Clark,
  Kenton Lee, and Luke Zettlemoyer. 2018.
\newblock \href {https://doi.org/10.18653/v1/N18-1202} {Deep contextualized
  word representations}.
\newblock In \emph{Proceedings of the 2018 Conference of the North {A}merican
  Chapter of the Association for Computational Linguistics: Human Language
  Technologies, Volume 1 (Long Papers)}, pages 2227--2237, New Orleans,
  Louisiana. Association for Computational Linguistics.

\bibitem[{Radford et~al.(2018)Radford, Narasimhan, Salimans, and
  Sutskever}]{radford2018improving}
Alec Radford, Karthik Narasimhan, Tim Salimans, and Ilya Sutskever. 2018.
\newblock Improving language understanding by generative pre-training.

\bibitem[{Shao et~al.(2018)Shao, Liu, Lai, Tseng, and Tsai}]{shao2018drcd}
Chih~Chieh Shao, Trois Liu, Yuting Lai, Yiying Tseng, and Sam Tsai. 2018.
\newblock Drcd: a chinese machine reading comprehension dataset.
\newblock \emph{arXiv preprint arXiv:1806.00920}.

\bibitem[{Sun et~al.(2019)Sun, Wang, Li, Feng, Chen, Zhang, Tian, Zhu, Tian,
  and Wu}]{sun2019ernie}
Yu~Sun, Shuohuan Wang, Yukun Li, Shikun Feng, Xuyi Chen, Han Zhang, Xin Tian,
  Danxiang Zhu, Hao Tian, and Hua Wu. 2019.
\newblock Ernie: Enhanced representation through knowledge integration.
\newblock \emph{arXiv preprint arXiv:1904.09223}.

\bibitem[{Tan and Zhang(2008)}]{tan2008empirical}
Songbo Tan and Jin Zhang. 2008.
\newblock An empirical study of sentiment analysis for chinese documents.
\newblock \emph{Expert Systems with applications}, 34(4):2622--2629.

\bibitem[{Vaswani et~al.(2017)Vaswani, Shazeer, Parmar, Uszkoreit, Jones,
  Gomez, Kaiser, and Polosukhin}]{vaswani2017attention}
Ashish Vaswani, Noam Shazeer, Niki Parmar, Jakob Uszkoreit, Llion Jones,
  Aidan~N Gomez, {\L}ukasz Kaiser, and Illia Polosukhin. 2017.
\newblock Attention is all you need.
\newblock In \emph{Advances in neural information processing systems}, pages
  5998--6008.

\bibitem[{Wu et~al.(2016)Wu, Schuster, Chen, Le, Norouzi, Macherey, Krikun,
  Cao, Gao, Macherey et~al.}]{wu2016google}
Yonghui Wu, Mike Schuster, Zhifeng Chen, Quoc~V Le, Mohammad Norouzi, Wolfgang
  Macherey, Maxim Krikun, Yuan Cao, Qin Gao, Klaus Macherey, et~al. 2016.
\newblock Google's neural machine translation system: Bridging the gap between
  human and machine translation.
\newblock \emph{arXiv preprint arXiv:1609.08144}.

\bibitem[{Xu et~al.(2020)Xu, Hu, Zhang, Li, Cao, Li, Xu, Sun, Yu, Yu, Tian,
  Dong, Liu, Shi, Cui, Li, Zeng, Wang, Xie, Li, Patterson, Tian, Zhang, Zhou,
  Liu, Zhao, Zhao, Yue, Zhang, Yang, Richardson, and Lan}]{clue}
Liang Xu, Hai Hu, Xuanwei Zhang, Lu~Li, Chenjie Cao, Yudong Li, Yechen Xu, Kai
  Sun, Dian Yu, Cong Yu, Yin Tian, Qianqian Dong, Weitang Liu, Bo~Shi, Yiming
  Cui, Junyi Li, Jun Zeng, Rongzhao Wang, Weijian Xie, Yanting Li, Yina
  Patterson, Zuoyu Tian, Yiwen Zhang, He~Zhou, Shaoweihua Liu, Zhe Zhao, Qipeng
  Zhao, Cong Yue, Xinrui Zhang, Zhengliang Yang, Kyle Richardson, and Zhenzhong
  Lan. 2020.
\newblock \href {https://doi.org/10.18653/v1/2020.coling-main.419} {{CLUE}: A
  {C}hinese language understanding evaluation benchmark}.
\newblock In \emph{Proceedings of the 28th International Conference on
  Computational Linguistics}, pages 4762--4772, Barcelona, Spain (Online).
  International Committee on Computational Linguistics.

\bibitem[{Xu et~al.(2021)Xu, Guo, Tang, Su, Shou, Gong, Zhong, Quan, Jiang, and
  Duan}]{xu-etal-2021-syntax}
Zenan Xu, Daya Guo, Duyu Tang, Qinliang Su, Linjun Shou, Ming Gong, Wanjun
  Zhong, Xiaojun Quan, Daxin Jiang, and Nan Duan. 2021.
\newblock \href {https://doi.org/10.18653/v1/2021.acl-long.420}
  {Syntax-enhanced pre-trained model}.
\newblock In \emph{Proceedings of the 59th Annual Meeting of the Association
  for Computational Linguistics and the 11th International Joint Conference on
  Natural Language Processing (Volume 1: Long Papers)}, pages 5412--5422,
  Online. Association for Computational Linguistics.

\bibitem[{Zhang et~al.(2022)Zhang, DOng, Wang, Wang, Wang, Liu, Huang, Li, and
  He}]{zhang2022revisiting}
Taolin Zhang, Junwei DOng, Jianing Wang, Chengyu Wang, Ang Wang, Yinghui Liu,
  Jun Huang, Yong Li, and Xiaofeng He. 2022.
\newblock Revisiting and advancing chinese natural language understanding with
  accelerated heterogeneous knowledge pre-training.
\newblock \emph{arXiv preprint arXiv:2210.05287}.

\bibitem[{Zhang et~al.(2021)Zhang, Zhang, Chen, Guo, Hua, Wang, and
  Zhou}]{zhang2021mengzi}
Zhuosheng Zhang, Hanqing Zhang, Keming Chen, Yuhang Guo, Jingyun Hua, Yulong
  Wang, and Ming Zhou. 2021.
\newblock Mengzi: Towards lightweight yet ingenious pre-trained models for
  chinese.
\newblock \emph{arXiv preprint arXiv:2110.06696}.

\bibitem[{Zhou et~al.(2020)Zhou, Zhang, Zhao, and Zhang}]{zhou-etal-2020-limit}
Junru Zhou, Zhuosheng Zhang, Hai Zhao, and Shuailiang Zhang. 2020.
\newblock \href {https://doi.org/10.18653/v1/2020.findings-emnlp.399}
  {{LIMIT}-{BERT} : Linguistics informed multi-task {BERT}}.
\newblock In \emph{Findings of the Association for Computational Linguistics:
  EMNLP 2020}, pages 4450--4461, Online. Association for Computational
  Linguistics.

\end{thebibliography}
\bibliographystyle{acl_natbib}

%\appendix

%\section{Appendix}
%\label{sec:appendix}

\end{document}